\pdfoutput=1

\documentclass[11pt]{article}

\usepackage[preprint]{acl}

\usepackage{times}
\usepackage{latexsym} 

\usepackage[T1]{fontenc}

\usepackage[utf8]{inputenc}

\usepackage{microtype}

\usepackage{inconsolata}

\usepackage{graphicx}

\usepackage{pifont}

\usepackage{placeins}
\usepackage{booktabs}
\usepackage{multirow}
\usepackage{arydshln}
\usepackage{graphicx}
\usepackage{paralist}
\usepackage{amssymb}
\usepackage{amsmath}
\usepackage{pifont}
\usepackage{xcolor}
\usepackage{bbm}
\usepackage{listings}
\usepackage{xspace}
\usepackage{algorithm}
\usepackage{algpseudocode}
\usepackage{colortbl}

\usepackage{cleveref}
\crefformat{section}{\S#2#1#3} 
\crefformat{subsection}{\S#2#1#3}
\crefformat{subsubsection}{\S#2#1#3}

\usepackage{algorithm}
\usepackage{algpseudocode}

\usepackage{pifont}

\usepackage{wrapfig,subcaption}

\definecolor{darkblue}{rgb}{0, 0, 0.5}
\hypersetup{colorlinks=true, citecolor=darkblue, linkcolor=darkblue, urlcolor=darkblue}

\title{Self-Routing RAG: Binding Selective Retrieval with \\ Knowledge Verbalization}

\author{
Di Wu\thanks{Equal Contribution}, Jia-Chen Gu\footnotemark[1], Kai-Wei Chang, Nanyun Peng\\
University of California, Los Angeles \\ 
\texttt{\{diwu,kwchang,violetpeng\}@cs.ucla.edu, gujc@ucla.edu}
}

\begin{document}
\maketitle

\begin{abstract}
Selective retrieval aims to make retrieval-augmented generation (RAG) more efficient and reliable by skipping retrieval when an LLM's parametric knowledge suffices. Despite promising results, existing methods are constrained by a binary design choice: either retrieve from a single external source or skip retrieval and let the LLM directly produce the final answer. We argue that this fallback underestimates the model's knowledge and obscures the more general multi-source decision problem that arises in practical systems. We propose \textbf{Self-Routing RAG (SR-RAG)}, which casts selective retrieval as knowledge source selection and treats the LLM itself as a first-class knowledge source. SR-RAG learns to select an appropriate knowledge source, optionally verbalize parametric knowledge, and answer using the selected source, all within a single left-to-right generation pass. SR-RAG further augments source selection by combining LLM-based uncertainty with a flexible external policy datastore to improve decision calibration. Across four benchmarks and three 7B-class LLMs, SR-RAG outperforms a strong selective retrieval baseline by 8.5\%/2.1\%/4.7\% while performing 26\%/40\%/21\% fewer retrievals, and it achieves favorable accuracy-latency trade-offs without dataset-specific threshold tuning.

\end{abstract}

\section{Introduction}

Retrieval-augmented generation (RAG) equips large language models (LLMs) with external knowledge at inference time, improving performance on tasks that require domain-specific or up-to-date knowledge \citep{DBLP:conf/iclr/KhandelwalLJZL20, DBLP:conf/nips/LewisPPPKGKLYR020, DBLP:conf/icml/BorgeaudMHCRM0L22, ram-etal-2023-context, shi-etal-2024-replug}. However, retrieval can be unnecessary when retrieved passages overlap with the LLM’s parametric knowledge, or even harmful when the passages are noisy or irrelevant. These limitations motivate \emph{selective retrieval}, where the system retrieves only when retrieval is likely to be useful \citep{he-etal-2021-efficient, mallen-etal-2023-trust, DBLP:conf/iclr/XuSC24, DBLP:conf/iclr/AsaiWWSH24, repoformer}.

\begin{figure*}[t]
\centering
\includegraphics[width=\textwidth]{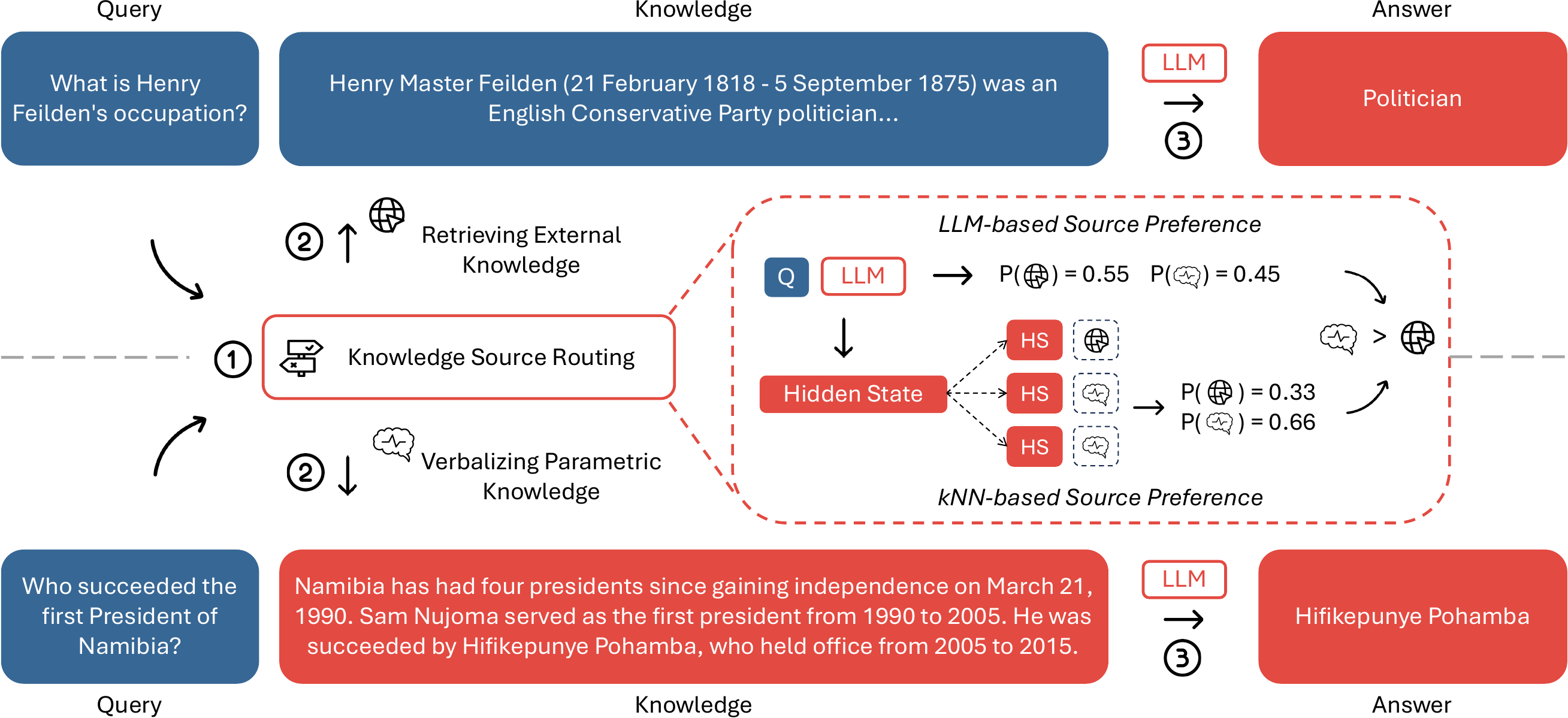}
\caption{\textbf{Self-Routing RAG Inference.} (1) A knowledge source is selected based on a query by combining the LLM's likelihood of a special source token with neighbor decisions from an external policy  datastore. (2) If an external source is selected, the passages are retrieved and appended as context. If the internal source is selected, the LLM generates a compact verbalized knowledge context. (3) The LLM then produces the final answer conditioned on the query and the selected knowledge. All steps can be executed in a single left-to-right generation pass.}
\label{fig:main-framework}
\end{figure*}

A limitation of current selective retrieval methods is reducing selective retrieval to a binary decision between consulting a single external source and a no-retrieval fallback where the LLM directly generates the final answer \citep{mallen-etal-2023-trust, jeong-etal-2024-adaptive, DBLP:conf/iclr/AsaiWWSH24, repoformer}. This design provides a coarse characterization of what the model can do without retrieval, as direct answer generation removes an opportunity to explicitly surface relevant parametric knowledge before committing to an answer. Consequently, routers trained by comparing RAG against direct answering, either via answer correctness \citep{wang-etal-2023-self-knowledge, repoformer} or likelihood \citep{he-etal-2021-efficient, DBLP:conf/iclr/XuSC24}, can learn preferences inconsistent with the LLM's non-retrieval capability. We observe this in a pilot study, where eliciting and conditioning on verbalized knowledge changes the preferred source for a substantial fraction of instances (\Cref{section-results-pilot-study}). Furthermore, the binary setup is inflexible as it cannot learn to leverage multiple knowledge sources, which could be an important limitation in real-world deployments.

We propose \textbf{Self-Routing RAG (SR-RAG)}, which addresses both limitations by casting selective retrieval as a \emph{knowledge source selection} problem and treating the LLM itself as a first-class knowledge source. SR-RAG expands the selective retrieval action space to routing between an internal knowledge source and an expandable set of external sources. As shown in \Cref{fig:main-framework}, at inference time, when the internal source is selected, the LLM first \emph{verbalizes} query-relevant parametric knowledge as a compact context, then answers conditioned on it. When an external source is selected, decoding is paused to retrieve evidence from the selected index, and the model answers conditioned on the retrieved passages.

To make SR-RAG reliable and efficient, we propose three techniques that span data construction, training, and inference. \textbf{(1) Verbalization-aware supervision:} given each training question, we collect multiple candidate contexts from each source, including diverse self-verbalized knowledge paragraphs, and assign the preferred source as the one whose contexts most increase the likelihood of the gold answer. This produces source selection labels that better reflect the LLM's capability without retrieval than direct-answer baselines. \textbf{(2) Multi-task learning:} we jointly train an LLM as a better knowledge source router, a better knowledge source, and a better knowledge reader in RAG. A two-staged multi-task objective is proposed that couples source selection, knowledge verbalization, and answer generation, including a preference alignment training to further boost the knowledge verbalization ability. \textbf{(3) kNN-augmented source selection:} purely relying on LLM-based uncertainty to select a source as done in previous work is brittle and agnostic to training-induced LLM ability shifts. We propose to build a policy datastore from light rollouts and combine the model's source selection probabilities with neighbor-defined source distributions in the hidden state space at test time.

We evaluate SR-RAG by fine-tuning Llama-2-7B-Chat \citep{touvron2023llama}, Phi-3.5-Mini-Instruct \citep{abdin2024phi}, and Qwen2.5-7B-Instruct \citep{yang2024qwen2} on a mixture of knowledge-intensive tasks and testing on four benchmarks. SR-RAG outperforms a strong standard selective retrieval baseline by 8.5\%/2.1\%/4.7\% while performing 26\%/40\%/21\% fewer retrievals across the three LLMs, respectively (\Cref{section-results-main}). We further show that SR-RAG makes more accurate source selection decisions (\Cref{section-results-selection-acc}) and achieves favorable accuracy-latency trade-offs without dataset-specific threshold tuning (\Cref{section-results-latency}, \Cref{fig:system-efficiency-main-text}). Ablations confirm that verbalization-aware labeling, preference alignment, and kNN-augmented inference each contribute materially to these gains (\Cref{section-results-ablation}). Finally, we experiment on routing among three sources (LLM, Wikipedia, and PubMed), where SR-RAG learns to retrieve from specialized corpus when needed while retaining strong performance on general workloads (\Cref{srrag-extensions}). Data and code will be publicly released at \url{https://github.com/xiaowu0162/self-routing-rag}.

\section{Related Work}
\label{section-related-work}

\paragraph{Selective Retrieval} To enhance the efficiency of RAG systems and reduce harmful or distracting retrievals, several works propose selectively skipping retrieval augmentation \citep{he-etal-2021-efficient, mallen-etal-2023-trust, DBLP:conf/iclr/XuSC24, repoformer}. A common approach assesses whether retrieval increases the likelihood of generating the correct answer and distills this signal into a supervised decision model \citep{he-etal-2021-efficient, schick2024toolformer, DBLP:conf/iclr/XuSC24}. Similarly, \citet{wang-etal-2023-self-knowledge} and \citet{repoformer} compare answers generated with and without retrieval to form supervision based on correctness. Other lines of work examine the question alone to decide whether retrieval is needed \citep{mallen-etal-2023-trust, jeong-etal-2024-adaptive, DBLP:conf/iclr/AsaiWWSH24}. To incorporate model confidence, recent methods use uncertainty-based selective retrieval \citep{DBLP:journals/corr/abs-2402-10612, DBLP:journals/corr/abs-2406-19215, moskvoretskii2025adaptiveretrievalselfknowledgebringing}. In contrast to prior work that treats skipping retrieval as direct answer generation, our work highlights that routing decisions and non-retrieval performance both improve when the LLM is treated as a knowledge source and is trained to verbalize query-relevant parametric knowledge.

\paragraph{Adaptive RAG Inference} This work also relates to adaptive RAG, which studies instance-specific inference strategies beyond a fixed retrieve-then-read pipeline. Prior work studies \emph{active retrieval}, where the system re-issues queries when initial retrieval is insufficient \citep{jiang-etal-2023-active, su-etal-2024-dragin}. Another line explores query decomposition and iterative retrieval to handle complex questions by breaking them into sub-queries \citep{shao-etal-2023-enhancing, kim-etal-2023-tree, liu-etal-2024-ra, lee-etal-2024-planrag}. Given retrieved results, \citet{DBLP:conf/iclr/AsaiWWSH24} and \citet{DBLP:journals/corr/abs-2401-15884} critique or revise the retrieved knowledge to improve output quality. \citet{parekh2024dynamic} incorporate an initial decision step to adaptively select the most suitable strategy based on the question. While SR-RAG focuses on the targeted decision of which knowledge source to consult, it introduces a complementary form of adaptivity by allowing the model to route among heterogeneous sources, including its own parametric knowledge, within a single generation pass.

\paragraph{LLMs as Knowledge Sources} A growing body of work explores using LLMs to generate auxiliary knowledge. Early studies show that LLMs can produce relevant background knowledge in zero-shot settings for commonsense reasoning \citep{shwartz-etal-2020-unsupervised, liu-etal-2022-generated}. More broadly, \citet{DBLP:conf/iclr/0002IWXJ000023} propose a generate-then-read approach, using the LLM as a context generator in place of external retrieval for RAG. LLMs have also been shown to generate effective intermediate reasoning steps for complex questions \citep{DBLP:conf/nips/Wei0SBIXCLZ22, DBLP:conf/nips/KojimaGRMI22}. Building on these insights, our work integrates LLM-generated knowledge into selective retrieval by making it an explicit, routable source, and by training the model to both decide when to rely on it and to produce compact verbalizations that improve end-to-end answering.

\section{Approach}
\label{section-approach}

We first reformulate selective retrieval as knowledge source selection. We then present Self-Routing RAG (SR-RAG), a framework that allows an LLM to self-route between external retrieval and internal knowledge verbalization within a single left-to-right generation pass.

\subsection{Problem Formulation}
\label{section-problem-formulation}

\paragraph{Knowledge Source-Aware Inference}
Given a user query $q$, a knowledge source $s$ returns relevant information as a text sequence $s(q)$ (possibly empty). An LLM reader $M$ then produces a response conditioned on both the query and the returned knowledge:
\begin{equation}
M(q, s(q)).
\end{equation}
Under this framework, standard RAG is an instantiation with $s$ being a retriever over an external datastore and $s(q)$ being retrieved passages.

\paragraph{Knowledge Source Selection}
In many practical settings, multiple knowledge sources may be available, denoted by
\begin{equation}
\mathcal{S} = \{\phi, s_1, \ldots, s_N\},
\end{equation}
where $\phi$ is a null source that always returns an empty string. A knowledge source selector $P$ chooses the most appropriate source for a given query, i.e., $P(q, \mathcal{S}) \in \mathcal{S}$. Combined with the downstream inference, the full pipeline becomes
\begin{equation}
M\big(q, P(q,\mathcal{S})(q)\big).
\end{equation}
Selective retrieval corresponds to the binary case $\mathcal{S}=\{\phi, s\}$, where the selector decides whether to invoke the retriever $s$ or to skip retrieval via $\phi$.

\subsection{Self-Routing RAG: Overview}
\label{section-method-overview}

We propose Self-Routing RAG (SR-RAG), which generalizes selective retrieval by treating the LLM itself as a first-class knowledge source. Concretely, SR-RAG routes between external sources $\{S_{e1}, ..., S_{em}\}$ (e.g., several external corpora) and an internal source $S_i$ (parametric knowledge). If $S_i$ is selected, the model does not directly answer. Instead, it first \emph{verbalizes} query-relevant parametric knowledge into a compact context and then answers conditioned on this context. This makes the non-retrieval path an explicit knowledge consultation step rather than an absence of context.

\begin{figure}[t!]
    \centering
    \includegraphics[scale=0.275]{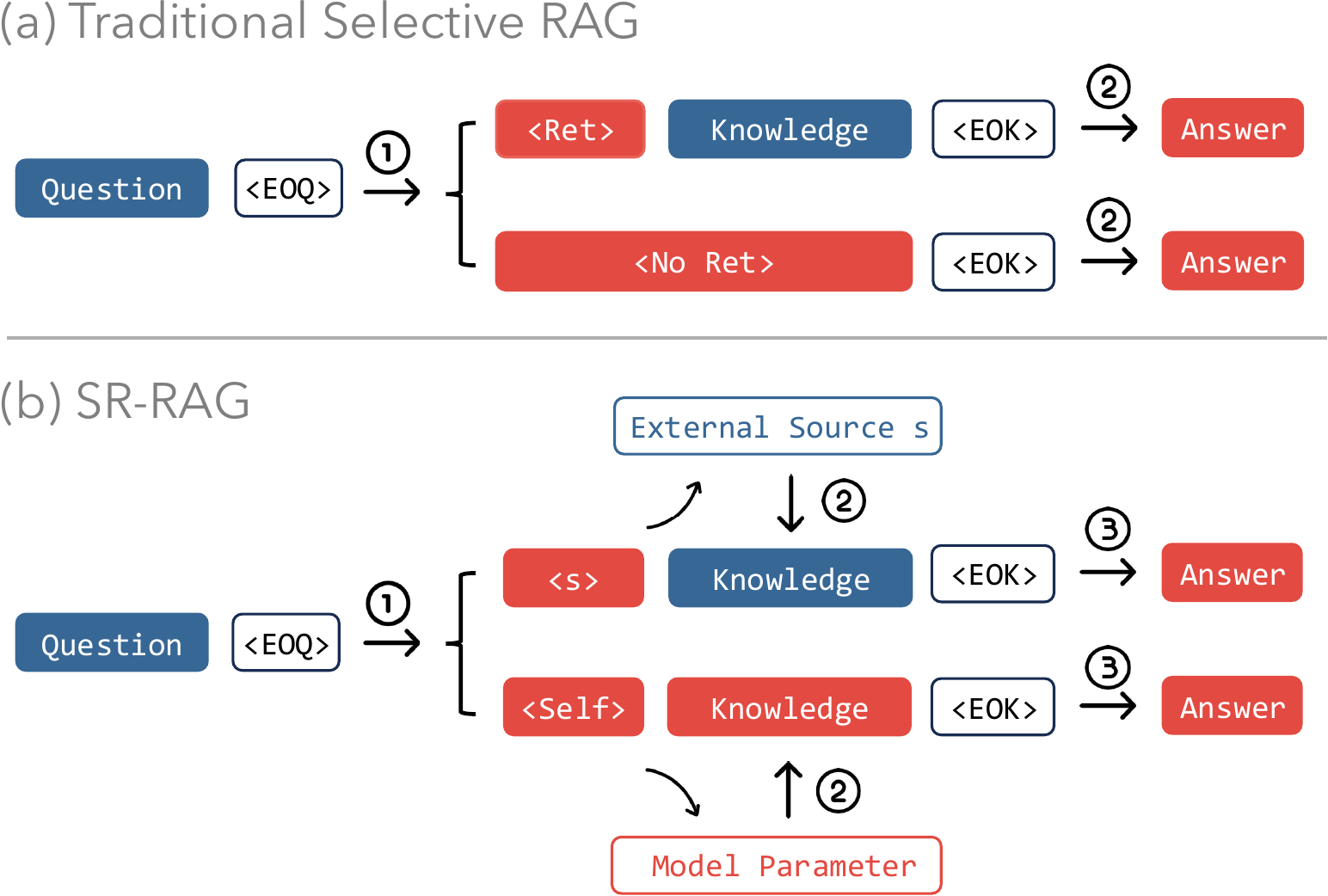}
    \caption{\textbf{SR-RAG expands selective retrieval into source routing.} Compared to traditional selective retrieval, SR-RAG allows the LLM to route among knowledge sources and to act as a knowledge source itself. Blue denotes external retrieved content and red denotes the LLM and its self-generated tokens.}
    \label{fig:inference-style}
\end{figure}

Following prior work that streamlines adaptive RAG inference via special-token prediction \citep{DBLP:conf/iclr/AsaiWWSH24, repoformer}, SR-RAG uses three sets of special tokens to implement routing in a single left-to-right generation pass:
\begin{compactenum}
    \item \texttt{<EOQ>}, which marks the end of the query and prompts source selection.\footnote{This design aligns with \citet{repoformer} but differs from \citet{DBLP:conf/iclr/AsaiWWSH24}. We find the additional \texttt{<EOQ>} helps the model allocate probability mass to source tokens without interfering with the query content.}
    \item A set of source tokens \texttt{<s>}, each representing a knowledge source $s\in\mathcal{S}$.\footnote{Our main setup uses a simple Wikipedia-based retrieval \texttt{<Wiki>} as the only $S_e$ and \texttt{<Self>} for $S_i$. This formulation naturally extends to more than two sources by adding additional source tokens. We study a three-source setting in \Cref{srrag-extensions}.}
    \item \texttt{<EOK>}, which indicates the end of the knowledge context and triggers answer generation.
\end{compactenum}
An illustration of the difference between SR-RAG and standard selective RAG is provided in \Cref{fig:inference-style}.

\subsection{Self-Routing RAG: Training}
\label{section-method-training}

We propose a fine-tuning recipe to improve any LLM's ability as the core model in the SR-RAG framework. Inspired by \citet{repoformer}, we propose to mine self-supervision from existing question-answer data $(q,a)$. 
Our pipeline uses the LLM itself as the internal source $S_i$ along with the external knowledge source, requiring neither human routing labels nor a stronger teacher model. For simplicity, we formulate with $\mathcal{S} = \{S_i, S_e\},$ here. The more general form readily follows and is formulated in \Cref{srrag-extensions}. 

\paragraph{Data Construction}
Our central goals are (1) labeling which source is more helpful for a given instance (2) collecting the most helpful knowledge when parametric knowledge is preferred. To do so, our rollout process first collects $n$ candidate contexts from each of the knowledge sources:
\begin{itemize}
    \item \emph{Parametric knowledge verbalization:} We use GenRead \citep{DBLP:conf/iclr/0002IWXJ000023} to elicit $n$ diverse verbalizations from the internal source $S_i$, denoted as $c_{i_1},\ldots,c_{i_n}$.
    \item \emph{External knowledge retrieval:} We retrieve $n$ context passages from the external source $S_e$, denoted as $c_{e_1},\ldots,c_{e_n}$.
\end{itemize}
Then, each candidate context $c_j$ is scored by the log-likelihood of generating the gold answer:
\begin{equation}
l_j = \log p_M(a \mid q, c_j).
\end{equation}
We then rank all $2n$ contexts by $l_j$ and select the preferred source $s\in\{S_i,S_e\}$ as the source that appears most frequently among the top-ranked contexts.
For convenience, we denote the highest- and lowest-scoring contexts from each source as $c_{i+}, c_{i-}, c_{e+}, c_{e-}$, respectively. The resulting training tuple is $(q,a,s,\{c_j,l_j\})$. The formal data collection algorithm is provided in \Cref{appendix-training-details}. 

\paragraph{Objective}
SR-RAG fine-tunes the LLM to perform (i) source selection, (ii) knowledge verbalization when $S_i$ is selected, and (iii) answer generation conditioned on the selected knowledge. We use a two-stage training procedure. 

\textbf{Stage 1} We perform behavior cloning by jointly optimizing three losses:
\begin{enumerate}
    \vspace{-1mm}
    \item $\mathcal{L}_{src}$, a cross-entropy loss that trains the model to predict the preferred source token immediately after \texttt{<EOQ>}:
    \begin{equation}
    \mathcal{L}_{src} = -\log p_M(\texttt{\textless s\textgreater}\mid q).
    \end{equation}
    \item $\mathcal{L}_{verb}$, a cross-entropy loss over the knowledge tokens when the LLM itself is labeled as the preferred knowledge source:
    \begin{equation}
    \mathcal{L}_{verb} =
    \begin{cases}
        -\log p_M(c_{i+}\mid q), & \text{if } s=S_i,\\
        0, & \text{if } s=S_e.
    \end{cases}
    \end{equation}
    \item $\mathcal{L}_{ans}$, a cross-entropy loss on generating the answer conditioned on the preferred knowledge:
    \begin{equation}
    \mathcal{L}_{ans} =
    \begin{cases}
        -\log p_M(a\mid q,c_{i+}), & \text{if } s=S_i,\\
        -\log p_M(a\mid q,c_{e+}), & \text{if } s=S_e.
    \end{cases}
    \end{equation}
\end{enumerate}
We combine these objectives as
\begin{equation}
\mathcal{L}_{stage1}=\mathcal{L}_{src}+\mathcal{L}_{verb}+\mathcal{L}_{ans}.
\end{equation}

\textbf{Stage 2} To further improve the knowledge verbalization quality, we apply Direct Preference Optimization (DPO, \citet{DBLP:conf/nips/RafailovSMMEF23}) using self-generated preference pairs $(c_{i+},c_{i-})$:
\begin{equation}
\mathcal{L}_{stage2}=\mathcal{L}_{src}+\mathcal{L}_{verb}^{DPO}+\mathcal{L}_{ans},
\end{equation}
\begin{equation}
\resizebox{\hsize}{!}{$
\mathcal{L}_{verb}^{DPO}=
\begin{cases}
-\log \sigma\Big( \beta \log \frac{p_M(c_{i+}\mid q)}{p_{ref}(c_{i+}\mid q)} - \beta \log \frac{p_M(c_{i-}\mid q)}{p_{ref}(c_{i-}\mid q)} \Big), & \text{if } s=S_i,\\
0, & \text{if } s=S_e.
\end{cases}
$}
\end{equation}
Here $M$ and $ref$ are initialized from the stage 1 model checkpoint, and only $M$ is updated. Conceptually, stage 2 distills a compute-intensive ``system 2'' process that searches over multiple verbalizations into a fast ``system 1'' behavior that produces a single high-quality knowledge verbalization during standard left-to-right decoding.

\subsection{Self-Routing RAG: Inference}
\label{section-method-inference}

As shown in \Cref{fig:main-framework,fig:inference-style}, SR-RAG inference proceeds in a single left-to-right generation pass with three steps: source selection, knowledge collection, and answer generation.

\paragraph{Nearest Neighbor-Enhanced Source Selection}
A common way to choose a source is to threshold the model probability of a source token, such as $p_M(\texttt{\textless Wiki\textgreater}\mid q)$ \citep{DBLP:conf/iclr/AsaiWWSH24, repoformer}. In practice, only relying on model's uncertainty can be brittle and does not take into account model ability changes after the fine-tuning. To make source selection more robust and controllable, we augment token likelihoods with a dynamic kNN policy signal derived from additional \emph{rollouts}. Concretely, we evaluate the fine-tuned model on a set of question-answer pairs\footnote{Empirically, we reuse the training set so that no additional data or supervision is required.}. For each pair, we compare the likelihood of generating the gold answer under different sources to decide a preferred source label. We then construct a \emph{policy datastore} that maps each query to its preferred source, using the hidden representation at \texttt{<EOQ>} as the key. At test time, we retrieve the $k$ nearest neighbors of the query in this representation space and form a neighbor-induced distribution over sources $p_D(\texttt{\textless s\textgreater}\mid q)$ from label counts. Finally, we select the source by applying a threshold on the combined score:
\begin{equation}
p_M(\texttt{\textless s\textgreater}\mid q)\times p_D(\texttt{\textless s\textgreater}\mid q).
\end{equation}
This combination adapts the decision boundary to the fine-tuned model's behavior while preserving the efficiency of a single decoding pass. It also improves SR-RAG's interpretability and controlability as the datastore consists of explicit source assignments, which can be audited, modified, or expanded by practitioners to steer routing behavior.

\paragraph{Knowledge Collection and Answer Generation}
After selecting a source, SR-RAG gathers knowledge and generates the final answer within the same left-to-right pass. If the internal source $S_i$ is selected, the model uses greedy decoding to verbalize a single compact knowledge context, which serves as a cost-efficient approximation to multi-sample verbalization. If an external source $S_e$ is selected, decoding is paused to retrieve passages from the chosen index, which are appended as the knowledge context. In both cases, the system appends \texttt{<EOK>} and the model generates the final answer conditioned on the query and the knowledge.

\section{Experimental Setup}

\subsection{Implementation Details of SR-RAG}
\label{section-implementation-details}

\paragraph{Data Construction} The main experiments are performed on two knowledge sources\footnote{We present multi-source extensions of SR-RAG in \Cref{srrag-extensions}.}: the 2018 English Wikipedia (\texttt{<Wiki>}) as the external knowledge source, and the LLM itself (\texttt{<Self>}) as the internal knowledge source. We use the official Wikipedia embeddings released by \citet{karpukhin-etal-2020-dense} and retrieve at a granularity of 100-word chunks. GenRead \citep{DBLP:conf/iclr/0002IWXJ000023} is used to verbalize diverse knowledge contexts. GenRead clusters zero-shot knowledge verbalizations in as in-context demonstrations to verbalize diverse knowledge. We limit the verbalized knowledge chunks to a maximum of 150 tokens. From each knowledge source, we collect $n$ = 5 knowledge contexts. 

\paragraph{Training} We fine-tune on a mixture of six short- and long-form knowledge-intensive datasets: Wizard of Wikipedia \citep{DBLP:conf/iclr/DinanRSFAW19}, Natural Questions \citep{kwiatkowski-etal-2019-natural}, FEVER \citep{thorne-etal-2018-fever}, OpenBookQA \citep{mihaylov-etal-2018-suit}, ARC-Easy \citep{DBLP:journals/corr/abs-2102-03315}, and ASQA \citep{stelmakh-etal-2022-asqa}. This mixture of 53,042 instances is a subset of the RAG instruction tuning data shown effective in \citet{DBLP:conf/iclr/AsaiWWSH24}. After running the data construction algorithm, 46.9\% of the instances are labeled with \texttt{<Self>} and the rest are labeled with \texttt{<Wiki>} as the preferred knowledge source, averaged across three LLMs. Full training details are provided in \Cref{appendix-training-details}.

\paragraph{Inference} To construct the policy datastore, we use the representation from a middle layer in the fine-tuned LLM\footnote{Layer 15 for Llama-2-7B-Chat and Phi-3.5–Mini-Instruct and layer 11 for Qwen2.5-7B-Instruct. We provide further visualizations and discussions of layer selection in \Cref{further-analyses-hidden-state-vis}.}, as middle layers are found to be indicative of LLM truthfulness \citep{DBLP:conf/icml/YinSC24}. At test time, the datastore index is cached on GPU and similarity search can be achieved via a single matrix multiplication. We retrieve $k$ = 30 nearest supporting examples and construct $p_D(\text{\texttt{\textless s\textgreater}} | q)$ from the counts of each knowledge source as the preferred source. Then, we impose a threshold $\tau$ on $p_M(\text{\texttt{\textless Wiki\textgreater}} | q)\times p_{D}(\text{\texttt{\textless Wiki\textgreater}} | q)$ to decide whether the external knowledge source should be selected.\footnote{$\tau$ = 0.1 for Llama-2-7B-Chat and $\tau$ = 0.2 for other models.}

\subsection{Evaluation}
\label{section-eval-details}
\paragraph{Datasets and Metrics} We evaluate on a diverse set of four knowledge-intensive NLP tasks. \textbf{PopQA} \citep{mallen-etal-2023-trust} is a free-form long-tail open-domain question answering dataset. Following \citet{DBLP:conf/iclr/AsaiWWSH24}, we use the subset of 1,399 questions that aims to test knowledge of long-tail entities. \textbf{TriviaQA} \citep{joshi-etal-2017-triviaqa} is an established open-domain question answering dataset that features relatively complex and diverse questions. We use the same test split and retrieval setup as in \citet{DBLP:conf/iclr/AsaiWWSH24}.  \textbf{PubHealth} \citep{DBLP:journals/corr/abs-2304-03728} is a fact-checking dataset focusing on the public health domain. Finally, \textbf{ARC Challenge}  \citep{DBLP:journals/corr/abs-2102-03315} is a multiple-choice question answering dataset featuring grade-school level science questions. Following common practice, we perform lexical post-processing of the model's output and report accuracy for PubHealth and ARC and substring matching for PopQA and TriviaQA.

\paragraph{Baselines} We compare SR-RAG with baselines that cover various training and inference strategies.
(1) First, using the LLM before fine-tuning, we compare with either always retrieving or always verbalizing with GenRead. (2) As illustrated in \Cref{fig:inference-style}, our main baseline is the state-of-the-art prior \textit{selective retrieval} pipeline, combining the ideas from \citet{he-etal-2021-efficient}, \citet{DBLP:conf/iclr/AsaiWWSH24}, and \citet{repoformer}. Specifically, the likelihoods of the LLM generating the answer with and without retrieval are used to create the knowledge source selection label. Then, we fine-tune the LLM for knowledge source selection (among $S_e$ and $\phi$) and generate the answer with optional retrieval. At inference, we apply a uniform threshold of 0.2 to the likelihood of the retrieval token for selective retrieval. (3) Always retrieving with the fine-tuned LLM after two-staged SR-RAG fine-tuning.

\section{Results}

\subsection{Knowledge Verbalization Alters Knowledge Source Preference}
\label{section-results-pilot-study}

\begin{figure*}[ht!]
\centering
\includegraphics[width=0.95\textwidth]{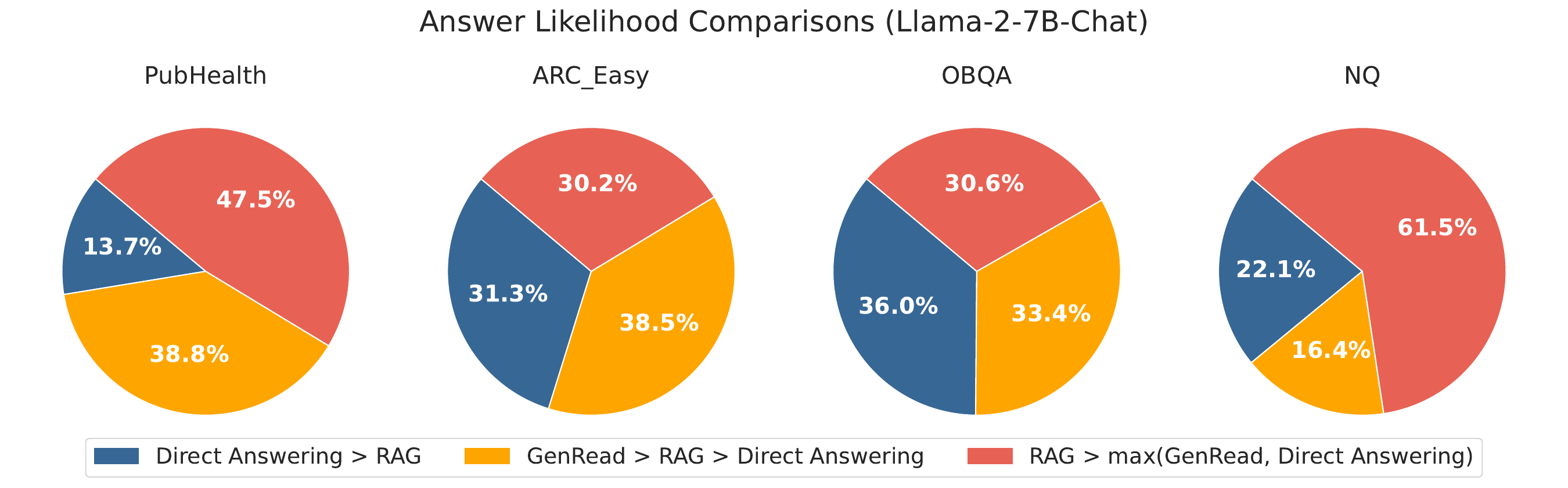}
\caption{\textbf{Knowledge verbalization significantly affects the LLM ability boundary.} For a large number of instances (16.4\% - 38.8\%, orange), GenRead reverses the knowledge source preferences: without considering GenRead, RAG dominates over parametric knowledge.}
\label{fig:pilot-likelihood-comparisons}
\end{figure*}
\begin{table*}[ht!]
\centering
\resizebox{\linewidth}{!}{
\def\arraystretch{1.1}
\begin{tabular}{c|c|cc|cc|cc|cc|cc}
\toprule
 \multirow{2}{*}{\textbf{Training}} & \multirow{2}{*}{\textbf{Inference}} & \multicolumn{2}{c|}{\textbf{PopQA}} & \multicolumn{2}{c|}{\textbf{TriviaQA}} & \multicolumn{2}{c|}{\textbf{PubHealth}} & \multicolumn{2}{c|}{\textbf{ARC}} & \multicolumn{2}{c}{\textbf{Average}} \\
 & &  \textbf{ACC} & \textbf{\%RAG} & \textbf{ACC} & \textbf{\%RAG} & \textbf{ACC} & \textbf{\%RAG} & \textbf{ACC} & \textbf{\%RAG} & \textbf{ACC} & \textbf{\%RAG} \\
\midrule
\rowcolor{gray!20} \multicolumn{12}{c}{\textbf{\texttt{Llama-2-7B-Chat}}} \\
\midrule
\multirow{2}{*}{No Fine-tuning} & Always RAG & 0.529 & 100\% & 0.641 & 100\% & 0.457 & 100\% & 0.546 & 100\% & 0.543 & 100\% \\
 & GenRead & 0.247 & 0\% & 0.616 & 0\% & 0.515 & 0\% & 0.605 & 0\% & 0.496 & 0\% \\
  \cdashline{1-12}
 \multirow{2}{*}{Selective RAG} & Always RAG & 0.567 & 100\% & 0.640 & 100\% & 0.588 & 100\% & 0.588 & 100\% & 0.596 & 100\% \\
 & Selective RAG & 0.565 & 98\% & 0.638 & 100\% & 0.589 & 100\% & 0.594 & 65\% & 0.597 & 86\% \\
 \cdashline{1-12}
 \multirow{2}{*}{SR-RAG} & Always RAG & \textbf{0.568} & 100\% & \textbf{0.669} & 100\% & 0.689 & 100\% & 0.608 & 100\% & 0.634 & 100\% \\
 & SR-RAG & 0.566 & 96\% & 0.664 & 89\% & \textbf{0.730} & 40\% & \textbf{0.630} & 29\% & \textbf{0.648} & 64\% \\
 \midrule
\rowcolor{gray!20} \multicolumn{12}{c}{\textbf{\texttt{Phi-3.5-Mini-Instruct}}} \\
\midrule
\multirow{2}{*}{No Fine-tuning} & Always RAG & 0.541 & 100\% & 0.594 & 100\% & 0.549 & 100\% & 0.771 & 100\% & 0.614 & 100\% \\
 & GenRead & 0.331 & 0\% & 0.567 & 0\% & 0.442 & 0\% & 0.840 & 0\% & 0.545 & 0\% \\
  \cdashline{1-12}
 \multirow{2}{*}{Selective RAG} & Always RAG & \textbf{0.570} & 100\% & 0.645 & 100\% & 0.701 & 100\% & 0.813 & 100\% & 0.682 & 100\% \\
 & Selective RAG & \textbf{0.570} & 100\% & 0.638 & 95\% & 0.704 & 91\% & 0.815 & 83\% & 0.682 & 92\% \\
 \cdashline{1-12}
 \multirow{2}{*}{SR-RAG} & Always RAG & 0.567 & 100\% & \textbf{0.659} & 100\% & 0.689 & 100\% & 0.820 & 100\% & 0.684 & 100\% \\
 & SR-RAG & 0.566 & 98\% & 0.657 & 92\% & \textbf{0.705} & 24\% & \textbf{0.854} & 5\% & \textbf{0.696} & 55\% \\
\midrule
\rowcolor{gray!20} \multicolumn{12}{c}{\textbf{\texttt{Qwen2.5-7B-Instruct}}} \\
\midrule
\multirow{2}{*}{No Fine-tuning} & Always RAG & 0.563 & 100\% & \textbf{0.667} & 100\% & 0.446 & 100\% & \textbf{0.916} & 100\% & 0.648 & 100\% \\
 & GenRead & 0.334 & 0\% & 0.626 & 0\% & 0.676 & 0\% & 0.875 & 0\% & 0.628 & 0\% \\
  \cdashline{1-12}
 \multirow{2}{*}{Selective RAG} & Always RAG & 0.555 & 100\% & 0.654 & 100\% & 0.600 & 100\% & 0.827 & 100\% & 0.659 & 100\% \\
 & Selective RAG & 0.529 & 88\% & 0.648 & 93\% & 0.608 & 82\% & 0.835 & 78\% & 0.655 & 85\% \\
 \cdashline{1-12}
 \multirow{2}{*}{SR-RAG} & Always RAG & \textbf{0.573} & 100\% & 0.662 & 100\% & 0.596 & 100\% & 0.821 & 100\% & 0.663 & 100\% \\
 & SR-RAG & 0.572 & 99\% & 0.659 & 89\% & \textbf{0.682} & 34\% & 0.830 & 46\% & \textbf{0.686} & 67\% \\
\bottomrule
\end{tabular}
}
\caption{\textbf{Main evaluation results on four tasks.} The best results are boldfaced. Across three backbone LLMs, SR-RAG consistently outperforms selective RAG and always retrieving while using a much lower retrieval budget. }
\label{tab:main-results}
\end{table*}

To motivate SR-RAG, we first show that knowledge verbalization can substantially change when an LLM benefits from retrieval. In a pilot study on Llama-2-7B-Chat using four datasets from our training mixture, we compare the likelihood of generating the gold answer under three conditions: no context (blue), the most helpful GenRead verbalization ($c_{i+}$, orange), and the most helpful retrieved passage ($c_{e+}$, red). As shown in \Cref{fig:pilot-likelihood-comparisons}, GenRead reverses the preferred knowledge source for a large fraction of instances, including 16\% of Natural Questions and over 30\% on the other datasets. This indicates that selective retrieval methods that omit verbalization can underestimate the LLM's effective capability without retrieval, leading to suboptimal source preference labels and weaker performance when retrieval is skipped. Additional details of the study setup are provided in \Cref{appendix-training-details}.

\begin{table}
    \centering
    \setlength{\tabcolsep}{3.2pt}
 \resizebox{\linewidth}{!}{
    \begin{tabular}{c|cccc|c}
    \toprule
     \textbf{Method} & \textbf{PopQA} & \textbf{TriviaQA} & \textbf{PubHealth} & \textbf{ARC} & \textbf{Average} \\
    \midrule
    \rowcolor{gray!20} \multicolumn{6}{c}{\textbf{Accuracy (Verbalization $\ge$ Retrieval)}} \\
    \midrule
     Self-RAG & 0.957 & 0.936 & 0.867 & 0.908 & 0.917 \\
    SR-RAG w/o. kNN & \textbf{0.959} & 0.930 & 0.869 & 0.888 & 0.912 \\
     SR-RAG & \textbf{0.959} & \textbf{0.943} & \textbf{0.880} & \textbf{0.910} & \textbf{0.923} \\
    \midrule
    \rowcolor{gray!20} \multicolumn{6}{c}{\textbf{AUROC (Retrieval $>$ Verbalization)}} \\
    \midrule
     Self-RAG & 0.489 & 0.503 & 0.438 & \textbf{0.557} & 0.497 \\
    SR-RAG w/o. kNN & 0.490 & \textbf{0.567} & 0.564 & 0.513 & 0.534 \\
     SR-RAG & \textbf{0.577} & 0.565 & \textbf{0.606 }& 0.533 & \textbf{0.570} \\
    \bottomrule
    \end{tabular}
    }
    \caption{\textbf{Source selection accuracy}. Llama-2-7B-Chat is used. SR-RAG achieves the best averaged performance in both evaluation settings.}
    \label{tab:decision-accuracy}
\end{table}

\begin{figure*}[t]{
\centering
\includegraphics[width=\linewidth]{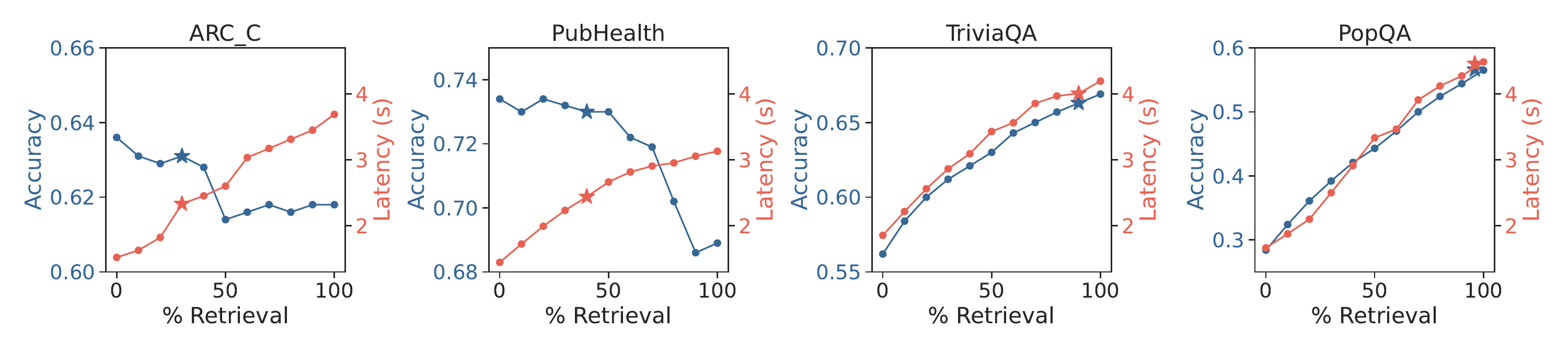}
\caption{\textbf{Accuracy and system latency of SR-RAG.} We show the performance of fine-tuned Llama-2-7B-Chat with different verbalization frequencies. SR-RAG's source selection policy (marked with stars) achieves near-optimal accuracy-efficiency trade-off without dataset-specific thresholds. }
\label{fig:system-efficiency-main-text}
}
\end{figure*}

\setlength{\tabcolsep}{2.7pt}
\begin{table*}[t]
\centering
\resizebox{\linewidth}{!}{
\def\arraystretch{1.1}
\begin{tabular}{c|c|cc|cc|cc|cc|cc}
\toprule
 \multirow{2}{*}{\textbf{Training}} & \multirow{2}{*}{\textbf{Inference}} & \multicolumn{2}{c|}{\textbf{PopQA}} & \multicolumn{2}{c|}{\textbf{TriviaQA}} & \multicolumn{2}{c|}{\textbf{PubHealth}} & \multicolumn{2}{c|}{\textbf{ARC}} & \multicolumn{2}{c}{\textbf{Average}} \\
 & &  \textbf{ACC} & \textbf{\%RAG} & \textbf{ACC} & \textbf{\%RAG} & \textbf{ACC} & \textbf{\%RAG} & \textbf{ACC} & \textbf{\%RAG} & \textbf{ACC} & \textbf{\%RAG} \\
\midrule
SR-RAG & SR-RAG & 0.566 & 96\% & \textbf{0.664} & 89\% & \textbf{0.730} & 40\% & \textbf{0.630} & 29\% & \textbf{0.648} & 64\% \\
 \midrule
 SR-RAG & SR-RAG w/o. kNN & 0.558 & 94\% & 0.658 & 77\% & 0.694 & 72\% & 0.627 & 56\% & 0.634 & 75\% \\
SR-RAG w/o. kv. label & SR-RAG & \textbf{0.568} & 100\% & 0.644 & 100\% & 0.598 & 100\% & 0.629 & 84\% & 0.610 & 96\% \\
 SR-RAG w/o. $\mathcal{L}_{verb}^{DPO}$ & SR-RAG & 0.564 & 98\% & 0.645 & 100\% & 0.674 & 100\% & 0.581 & 66\% & 0.616 & 86\% \\
 \bottomrule
\end{tabular}
}
\caption{\textbf{Ablation studies on SR-RAG.} Llama-2-7B-Chat is used as the LLM.}
\vspace{-2mm}
\label{tab:ablation-study}
\end{table*}

\subsection{SR-RAG Improves Accuracy with Reduced Retrieval Frequency}
\label{section-results-main}

\Cref{tab:main-results} shows the end-to-end generation performance on three LLMs, demonstrating the advantage of SR-RAG over always retrieving, selective retrieval, and other baselines. Relative to always retrieving with the pretrained model, SR-RAG improves accuracy by 19.3\% (Llama-2-7B-Chat), 13.4\% (Phi-3.5-Mini-Instruct), and 5.9\% (Qwen2.5-7B-Instruct). Compared to the selective retrieval baseline, SR-RAG achieves 8.5\%/2.1\%/4.7\% higher accuracy while triggering 26\%/40\%/21\% fewer retrievals across the three models, respectively. Notably, the selective retrieval baseline saves only 8--15\% retrievals and achieves nearly identical accuracy to always retrieving, suggesting that skipping retrieval alone is not sufficient to reliably identify queries that can be handled without external context. In contrast, SR-RAG reduces retrieval by 20--40\% while maintaining or improving accuracy, driven by accurate source selection and a stronger non-retrieval path.

Remarkably, without threshold tuning, SR-RAG can \textit{dynamically adapt} its retrieval behavior.  For PopQA, which emphasizes long-tail facts, SR-RAG retrieves external knowledge more frequently. For PubHealth and ARC, where the model can often rely on internal knowledge, SR-RAG skips retrieval more and can outperform always retrieving. We find that this flexibility also extends to selecting a specialized external source when multiple are present (\Cref{srrag-extensions}). We also present qualitative examples in the appendix (\Cref{qualitative-study-1,qualitative-study-2}).

\subsection{SR-RAG Makes Accurate Source Selection Decisions}
\label{section-results-selection-acc}

Can SR-RAG choose the appropriate knowledge source? \Cref{tab:decision-accuracy} compares SR-RAG with Self-RAG \citep{DBLP:conf/iclr/AsaiWWSH24} and an SR-RAG variant without kNN augmentation. We evaluate decision quality under two criteria: (1) abstaining from retrieval when skipping retrieval does not harm performance and (2) retrieving only when retrieval yields strictly better context. Under the first criterion, all methods perform well, and SR-RAG achieves the best overall accuracy. Under the stricter second criterion, SR-RAG substantially improves AUROC, outperforming Self-RAG by 14.7\% on average. Removing kNN augmentation leads to large drops in both accuracy and AUROC, highlighting the benefit of adapting source selection to fine-tuning-induced shifts in model behavior. We visualize the hidden state space in \Cref{further-analyses-hidden-state-vis} to provide further intuitions.

\subsection{System Efficiency}
\label{section-results-latency}

We measure end-to-end latency using Llama-2-7B-Chat under a batched inference setting (details in \Cref{appendix-latency-formulation}). \Cref{fig:system-efficiency-main-text} shows the accuracy-latency trade-off as we vary retrieval frequency. As expected, latency improves as fewer instances invoke retrieval, and the optimal retrieval rate varies across datasets. Importantly, SR-RAG's learned policy achieves a near-optimal accuracy-efficiency trade-off across datasets without dataset-specific threshold tuning. These results indicate that SR-RAG improves both accuracy and efficiency in a robust manner.

\subsection{Ablation Study}
\label{section-results-ablation}

We ablate SR-RAG components on Llama-2-7B-Chat in \Cref{tab:ablation-study}. Disabling kNN-based routing (w/o.\ kNN) increases retrieval frequency and reduces accuracy, confirming that kNN augmentation improves robustness to ability shifts introduced by fine-tuning. Removing knowledge verbalization during preference labeling (w/o.\ kv.\ label) causes the model to over-rely on retrieval and degrades performance, showing that verbalization-aware supervision is important for accurate routing. Finally, ablating stage-2 preference alignment (training stage 2 with the stage-1 objective) reduces verbalization quality and leads to substantially more retrievals, consistent with weaker internal knowledge generation. Together, these results indicate that all three components contribute materially to SR-RAG's overall gains. We provide additional analyses in \Cref{appendix-further-analyses} including the choice for source labeling heuristics, the size of the policy datastore, and other hyperparameters.

\section{Conclusion}
We present SR-RAG, a novel retrieval-augmented generation (RAG) framework that tightly integrates selective retrieval with knowledge verbalization. By reformulating selective retrieval as a knowledge source selection problem, SR-RAG enables the LLM to not only choose between external and internal knowledge sources but also to serve as a knowledge source itself. During inference, SR-RAG leverages internal hidden states and a nearest-neighbor policy to make accurate, adaptive source selection decisions. Extensive experiments show that SR-RAG significantly improves answer accuracy while reducing retrieval frequency and latency, offering a scalable and reliable path forward for more efficient, knowledge-aware RAG systems.

\bibliography{custom}

\clearpage
\appendix
\twocolumn[{%
 \centering
 \Large\bf Supplementary Material: Appendices \\ [20pt]
}]

\label{sec:appendix}

\section{SR-RAG: Further Details}

\subsection{List of Notations}

In \Cref{tab:notations}, we present the major notations and parameters used throughout the paper. 

\begin{table}[h]
    \centering
    \resizebox{\linewidth}{!}{
    \begin{tabular}{cl}
        \toprule
        \textbf{Notation} & \textbf{Description} \\
        \midrule
        $q$  & User query input to the system. \\
        $a$  & The expected answer. \\
        $M$  & The LLM. \\
        $\mathcal{S}$  & Set of all knowledge sources. \\
        $S_e$  & The external knowledge source. \\
        $S_i$  & The internal knowledge source (parametric knowledge). \\
        $c_{i+}$  & Most helpful verbalized knowledge context from $S_i$. \\
        $c_{i-}$  & Least helpful verbalized knowledge context from $S_i$. \\
        $c_{e+}$  & Most helpful retrieved knowledge context from $S_e$. \\
        $c_{e-}$  & Least helpful retrieved knowledge context from $S_e$. \\
        \texttt{<EOQ>} & End-of-query special token. \\
        \texttt{<EOK>} & End-of-knowledge special token. \\
        \texttt{<s>} & Special token representing knowledge source $s\in\mathcal{S}$. \\
        \texttt{<Wiki>} & Special token representing Wikipedia. \\
        \texttt{<Self>} & Special token representing the LLM itself as knowledge source. \\
        $k$ & number of neighbors retrieved for source policy inference \\
        \bottomrule
    \end{tabular}
    }
    \caption{\textbf{A summary of the key symbols.}}
    \label{tab:notations}
\end{table}

\subsection{Training Details}
\label{appendix-training-details}

\paragraph{Dataset Construction Algorithm} \Cref{algo-srrag-data} presents the full algorithm for constructing training data and labeling knowledge source preferences. GenRead is executed independently on each training data subset. We adopt instance-level notation for clarity. The pipeline naturally scales to additional knowledge sources by applying knowledge collection and likelihood evaluation in parallel across sources.

\begin{algorithm}[h]
\caption{\small\textbf{SR-RAG Training Data Construction}}
\small
\label{algo-srrag-data}
\begin{algorithmic}[1]
\Require LLM $M$, External Retriever $\mathcal{R}$, Dataset $\mathcal{D}$, Number of contexts $n$
\For{$(q, a) \in \mathcal{D}$}
    \State // Retrieving External Knowledge
    \State $\mathcal{C}_{S_e} \gets \mathcal{R}(q, n)$ 
    \State // Knowledge Verbalization
    \State $\mathcal{C}_{S_i} \gets$ $M$.GenRead$(q, n)$
    
    \State // Compute Likelihoods
    \For{$c \in \mathcal{C}_{S_e} \cup \mathcal{C}_{S_i}$}
        \State $l_{c} \gets p_M(a | q, c)$ 
    \EndFor

    
    \State $s \gets \arg\max_{s \in \{S_e, S_i\}} \sum_{c \in \mathcal{C}_s} l_{c}$
    \State Store $(q, a, s, \{c, l_{c}\})$
\EndFor

\State \Return Processed dataset with labeled knowledge sources
\end{algorithmic}
\end{algorithm}

\paragraph{GenRead Prompt} We implement GenRead following the setup in \citet{DBLP:conf/iclr/0002IWXJ000023}. In the second verbalization round, five clusters with five in-context examples each are used. All datasets except ASQA follow the general prompt shown in \Cref{genread-prompt-general}. For ASQA, we include an additional instruction to handle ambiguity: “\texttt{If the question is ambiguous, generate multiple documents for each possibility.}”

\begin{figure}[ht]
    \centering
    \small
    \fbox{\begin{tabular}{ p{0.965\linewidth} }
    \texttt{Generate a background document from Wikipedia to help answer the following question. Directly start with document content and do not generate URL.
    }
    \\
    \\
    \texttt{Question: \{question\}} \\
    \\
    \texttt{Background document:}
    \end{tabular}}
    \caption{\textbf{Prompt used for knowledge verbalization data collection via GenRead.}}
    \label{genread-prompt-general}
\end{figure}

\paragraph{Training Data} \Cref{tab:data_stats} summarizes the training and validation splits, including the proportion of examples where self-verbalized knowledge is preferred over retrieved knowledge.

\begin{table}[h]
    \centering
    \resizebox{\linewidth}{!}{
    \begin{tabular}{l|ccc|ccc}
        \toprule
        \multirow{2}{*}{\textbf{Dataset}} & \multirow{2}{*}{\textbf{Train}} & \multirow{2}{*}{\textbf{Validation}} & \multirow{2}{*}{\textbf{Total}} & \multicolumn{3}{c}{\textbf{\%Verbalization}} \\
         & & & & \textbf{Llama} & \textbf{Phi} & \textbf{Qwen} \\
        \midrule
        ARC\_Easy & 2037 & 107 & 2144 & 61\% & 84\% & 66\% \\
        NQ & 14753 & 776 & 15529 & 28\% & 33\% & 41\% \\
        OBQA & 4462 & 234 & 4696 & 61\% & 77\% & 61\% \\
        FEVER & 9467 & 498 & 9965 & 52\% & 58\% & 68\% \\
        WoW & 16493 & 868 & 17361 & 13\% & 55\% & 32\% \\
        ASQA & 3700 & 194 & 3894 & 13\% & 25\% & 16\% \\
        \bottomrule
    \end{tabular}
    }
    \caption{\textbf{Statistics of the training and validation data with verbalization percentages.} Llama = Llama-2-7B-Chat, Phi = Phi-3.5-Mini-Instruct, and Qwen = Qwen2.5-7B-Instruct.}
    \label{tab:data_stats}
\end{table}

\paragraph{Training Process} To fully leverage the backbone LLM's ability to follow natural language instructions, both SR-RAG fine-tuning and inference use the following prompt that interleaves special tokens with natural language:
\begin{center}
    \texttt{Question: \{question\} Background knowledge: <EOQ> \\ <s> \{knowledge\} <EOK> Answer: \{answer\}}
\end{center} 
During training, the loss is computed on the knowledge part only if \texttt{<s>} is \texttt{<Self>}. If \texttt{<s>} is \texttt{<Wiki>}, we augment $c_{e+}$ by randomly appending $\max(p-1, 0)$ retrieved contexts, where $p$ is sampled from a Poisson distribution with $\lambda=2$. This data augmentation improves the LLM's robustness to different retrieval strategies and various levels of retrieval quality. For stage 1 training, we use batch size 64, learning rate 1e-5, and fine-tune for 1 epoch. For stage 2 training, we use batch size 64, learning rate 5e-7, $\beta$ = 0.3 for DPO, and train for another epoch. All the experiments are performed on a machine with eight A800 (80GB) GPUs and a machine with eight A6000 GPUs. On eight A800 (80GB) GPUs, the two-stage training takes approximately 10 hours for a 7B-sized model. 

\subsection{Latency Formulation}
\label{appendix-latency-formulation}

To evaluate the inference efficiency of SR-RAG, we measure the latency in a realistic batched inference setup, where the system handles a batch of $B$ = 10 queries and returns the results for all queries. For the latency experiments presented in the paper, we decompose the system latency as follows:

\begin{itemize}
    \item \textbf{Source Selection Time} ($T_d$): The time for the knowledge source selector to determine whether to retrieve from external sources or rely on parametric knowledge.
    \item \textbf{Retrieval Latency} ($T_{rs}$): The time for fetching external knowledge from the database if the model chooses to retrieval from an external knowledge source $s$. In our batched setting, we calculate $T_{rs}$ by performing a batched retrieval for all the $B$ instances that require retrieval and report the per-item latency.
    \item \textbf{Verbalization Latency} ($T_v$): The time for the LLM to verbalize parametric knowledge if the LLM itself is selected as the knowledge source.
    \item \textbf{Generation Latency} ($T_g$): The time for the LLM to generate the response, conditioned on either retrieved or verbalized knowledge.
\end{itemize}

Thus, the total per-item latency $T_{\text{total}}$ is given by:

\[
T_{\text{total}} =
\begin{cases}
T_d + T_v + T_g, & \text{if verbalize}, \\
T_d + T_{rs} + T_g, & \text{if retrieve from $s$}.
\end{cases}
\]

We have the following remarks:
\begin{itemize}
    \item This formulation assumes that both the retrieval index and the source selection datastore are pre-constructed. This assumption is reasonable as these indices are only constructed once and then constantly reused.
    \item We choose the batched setup due to the complexity of the retrieval system. For instance, in our implementation of Wikipedia search, it takes around five to ten seconds per instance to encode the query and retrieve the most relevant context chunks. The batched setting amortizes the latency of single retrieval. 
    \item As the nearest-neighbor search only involves one matrix product for similarity calculation and one top-$k$ operation, $T_d$ is generally very small. In fact, we find $T_d$ (0.01s) $<<$ $T_g$ (0.1s) $<$ $T_v$ (1s). On the other hand, $T_{rs}$ is the major bottleneck of the pipeline. As a result, in an online setting, the system's efficiency gain directly converges to the percentage of retrievals it is able to avoid.
\end{itemize}

\begin{table*}[ht!]
\centering
\resizebox{0.95\linewidth}{!}{
\def\arraystretch{1.1}
\begin{tabular}{c|c|c|c|c|c|c|c}
\toprule
\multirow{1}{*}{\textbf{Training}} & \multirow{1}{*}{\textbf{Inference}} & \textbf{PubMedQA} & \textbf{PopQA} & \textbf{TriviaQA} & \textbf{PubHealth} & \textbf{ARC} & \textbf{Average} \\
\midrule
\rowcolor{gray!20} \multicolumn{8}{c}{\textbf{\texttt{Llama-2-7B-Chat}}} \\
\midrule
\multirow{3}{*}{No Fine-tuning}
 & GenRead & 0.544 & 0.247 & 0.616 & 0.515 & 0.605 & 0.505 \\
 & Always RAG, Wiki & 0.636 & 0.529 & 0.641 & 0.457 & 0.546 & 0.562 \\
 & Always RAG, PubMed & 0.656 & 0.157 & 0.470 & 0.612 & 0.537 & 0.486 \\
 \cdashline{1-8}
\multirow{7}{*}{SR-RAG  (3-source)}
 & Always RAG, Wiki & 0.342 & \textbf{0.587} & 0.639 & 0.668 & 0.639 & 0.572 \\
 & Always RAG, PubMed & 0.672 & 0.325 & 0.539 & 0.713 & 0.645 & 0.642 \\
 & Always Verbalizing & 0.388 & 0.305 & 0.572 & \textbf{0.724} & 0.631 & 0.579 \\
 & SR-RAG & \textbf{0.680} & 0.583 & \textbf{0.647} & \textbf{0.724} & \textbf{0.655} & \textbf{0.677} \\
 & \multicolumn{1}{r|}{(\% \texttt{<Wiki>})} & 6\% & 95\% & 88\% & 17\% & 4\% & 29\% \\
 & \multicolumn{1}{r|}{(\% \texttt{<Pubmed>})} & 91\% & 4\% & 7\% & 70\% & 86\% & 64\% \\
 & \multicolumn{1}{r|}{(\% \texttt{<Self>})} & 3\% & 1\% & 5\% & 13\% & 10\% & 8\% \\
\midrule
\rowcolor{gray!20} \multicolumn{8}{c}{\textbf{\texttt{Phi-3.5-Mini-Instruct}}} \\
\midrule
\multirow{3}{*}{No Fine-tuning}
 & GenRead & 0.376 & 0.331 & 0.567 & 0.442 & \textbf{0.840} & 0.511 \\
 & Always RAG, Wiki & 0.460 & 0.541 & 0.594 & 0.549 & 0.771 & 0.583 \\
 & Always RAG, PubMed & 0.676 & 0.143 & 0.356 & 0.447 & 0.726 & 0.470 \\
 \cdashline{1-8}
\multirow{7}{*}{SR-RAG  (3-source)}
 & Always RAG, Wiki & 0.486 & \textbf{0.580} & 0.634 & 0.676 & 0.805 & 0.636 \\
 & Always RAG, PubMed & 0.738 & 0.302 & 0.494 & 0.644 & 0.813 & 0.598 \\
 & Always Verbalizing & 0.498 & 0.327 & 0.510 & 0.704 & 0.819 & 0.572 \\
 & SR-RAG & \textbf{0.736} & 0.560 & \textbf{0.639} & \textbf{0.716} & 0.824 & \textbf{0.695} \\
 & \multicolumn{1}{r|}{(\% \texttt{<Wiki>})} & 1\% & 95\% & 87\% & 54\% & 0\% & 47\% \\
 & \multicolumn{1}{r|}{(\% \texttt{<Pubmed>})} & 97\% & 3\% & 10\% & 18\% & 27\% & 31\% \\
 & \multicolumn{1}{r|}{(\% \texttt{<Self>})} & 2\% & 2\% & 3\% & 28\% & 73\% & 22\% \\
\midrule
\rowcolor{gray!20} \multicolumn{8}{c}{\textbf{\texttt{Qwen2.5-7B-Instruct}}} \\
\midrule
\multirow{3}{*}{No Fine-tuning}
 & GenRead & 0.418 & 0.334 & 0.626 & 0.676 & 0.875 & 0.586 \\
 & Always RAG, Wiki & 0.612 & 0.563 & \textbf{0.667} & 0.446 & 0.916 & 0.641 \\
 & Always RAG, PubMed & 0.698 & 0.129 & 0.469 & 0.401 & 0.821 & 0.504 \\
 \cdashline{1-8}
\multirow{7}{*}{SR-RAG  (3-source)}
 & Always RAG, Wiki & 0.358 & \textbf{0.576} & 0.639 & 0.637 & 0.818 & 0.606 \\
 & Always RAG, PubMed & 0.726 & 0.307 & 0.503 & 0.676 & 0.824 & 0.607 \\
 & Always Verbalizing & 0.522 & 0.303 & 0.550 & \textbf{0.695} & 0.828 & 0.580 \\
 & SR-RAG & \textbf{0.730} & 0.572 & 0.651 & 0.681 & \textbf{0.834} & \textbf{0.694} \\
 & \multicolumn{1}{r|}{(\% \texttt{<Wiki>})} & 15\% & 96\% & 86\% & 20\% & 0\% & 43\% \\
 & \multicolumn{1}{r|}{(\% \texttt{<Pubmed>})} & 79\% & 1\% & 13\% & 48\% & 39\% & 36\% \\
 & \multicolumn{1}{r|}{(\% \texttt{<Self>})} & 6\% & 3\% & 1\% & 32\% & 61\% & 21\% \\
\bottomrule
\end{tabular}
}
\caption{\textbf{Three-source results.} We route between retrieving from \textbf{Wikipedia}, retrieving from \textbf{PubMed}, or \textbf{self-verbalizing} knowledge. For each model we report task accuracy and the selection frequencies of the individual sources \texttt{wiki/PubMed/self} under SR-RAG's LLM+kNN knowledge source routing scheme.}
\label{tab:3way-routing-results}
\end{table*}

\section{SR-RAG: Multi-Source Extensions}
\label{srrag-extensions}

Beyond routing between the LLM itself and an external knowledge source, SR-RAG's formulation is naturally compatible with routing between multiple external knowledge sources as well. In this section, we provide further experiments on routing between three knowledge sources and discuss other potential multi-source applications.

\subsection{Experiment: Routing between General and Specialized Sources}

We perform experiments on the PubMed corpus \citep{ncbi-pubmed}, which contains professional biomedical literature. Compared to Wikipedia and the LLM itself, PubMed has the unique advantage of featuring long-tail specialized biomedical knowledge. Meanwhile, it falls short at providing knowledge for many general factuality questions, posing a challenge to accurately routing the knowledge source. 

Based on this new setup, we formulate SR-RAG with three knowledge source tokens: \texttt{<Self>}, \texttt{<Wiki>}, and \texttt{<Pubmed>}. We leverage the MedRAG toolkit \citep{xiong-etal-2024-benchmarking} to implement retrieval on PubMed. The training and inference strategies of the two-source SR-RAG recipe naturally extends to this formulation and are followed in our experiments here. For training and evaluation, we additionally use PubMedQA \citep{jin-etal-2019-pubmedqa}, a widely used biomedical question answering dataset. PubMedQA features short questions extracted from PubMed abstracts and provides associated ground truth yes/no/maybe answers. We sample 10k artificially generated documents for training (combined with the data mixture used in the main text) and 500 examples from the expert-labeled subset for evaluation. The same set of LLMs from three families (Llama, Qwen, Phi) are trained and evaluated. 

In \Cref{tab:3way-routing-results}, we present the three-way routing results of SR-RAG and compare with always-RAG as well as GenRead with the model before fine-tuning. Our main findings are as follows:

\begin{itemize}
    \item \textbf{Adding PubMed as an additional knowledge source does not harm its effectiveness.} The results on datasets other than PubMedQA align with the performance of two-source SR-RAG reported in the main text.
    \item \textbf{SR-RAG training improves the model's ability to verbalize its knowledge} on PubMedQA as well while being much more efficient than GenRead, as shown by the better performance of SR-RAG + Always Verbalizing compared to GenRead. 
    \item \textbf{SR-RAG intelligently routes the best knowledge source for different workloads.} For PubMedQA, PubHealth, and ARC, where PubMed can outperform Wiki by a large margin, SR-RAG frequently self-selects PubMed as the knowledge source. On the other hand, on general factuality benchmarks such as PopQA and TriviaQA, SR-RAG selects Wikipedia more frequently. More importantly, SR-RAG achieves this behavior end-to-end without the need for any manual interventions.
    \item \textbf{On PubMedQA, the end-to-end SR-RAG achieves the best performance compared to the baselines.} The trend is consistent across three different LLMs. 
\end{itemize}

Overall, these experiments confirm the effectiveness of extending SR-RAG to three sources and suggests a promising outlook for further extension.

\subsection{Discussion: Further Applications}

SR-RAG has numerous useful real-world applications and exhausitively experimenting on them is beyond the scope of this paper's experiments. In this section, we discuss several potential applications that SR-RAG naturally supports. 

\paragraph{Cost-Aware Routing} Querying different knowledge sources brings distinct latency costs. SR-RAG's preference label can be constructed such that the cheapest useful knowledge source is preferred, improving the latency while preventing performance degradation. In addition, varied levels of compute are required to verbalize the required knowledge for different types of queries. SR-RAG can be used to model this pattern by formulating multiple internal knowledge sources associated with different levels of compute. This philosophy connects with the literature on adaptive reasoning budget \citep{lou2025adacot}.  

\paragraph{Recency-Aware Routing} Many user queries have an inherent temporal dimension (e.g., “today,” “as of Sept. 2025,” fast-moving entities) where stale knowledge degrades correctness. SR-RAG can expose “freshness-tiered” sources and learn a preference label that prioritizes the most up-to-date tier when temporal cues or out-of-vocabulary entities are detected. The router can condition on time expressions, named-entity recency features, and source timestamps. When the model’s self-estimate is uncertain or predates the query’s time scope, it escalates to a recency-guaranteed index. 

\paragraph{Privacy-Aware Routing} Queries often mix public facts with sensitive tenant data (e.g., customer tickets, internal codenames, emails). SR-RAG can represent privacy tiers as distinct sources and train preference labels under a “least-privilege first” policy: prefer private indices when identifiers or confidentiality cues are present and fall back to public only when the query is non-sensitive and private stores are irrelevant.

\section{Further Analyses}
\label{appendix-further-analyses}

\subsection{Self-Routing RAG: Hyperparameters}
\label{further-analyses-srrag-hyperparams}

In this section, we analyze other hyperparameters in SR-RAG. 

\paragraph{Source Labeling Heuristics} In SR-RAG, the default design for source preference labeling is to collect the knowledge from each source and to select the top-ranked source among the top 50\% of knowledge contexts in terms of contribution to the likelihood of the answer. In this section, we compare this strategy to a number of alternative heuristics:

\begin{itemize}
    \item \textbf{Best Single Likelihood:} Selecting the source that produces the knowledge leading to the highest answer likelihood. 
    \item \textbf{Best Average Likelihood:} Selecting the source that leads to the highest answer likelihood, averaged over all knowledge contexts.
    \item \textbf{Best All Rank:} Selecting top-ranked knowledge source using all the knowledge context instead of top-50\%.
\end{itemize}

In \Cref{tab:source_lab_heuristics}, we present the F1 score and AUROC using Llama-2-7B-Chat as the LLM. Correctness is defined as preferring self-verbalized knowledge when it is better than or equal to performing retrieval in terms of the downstream question answering performance. Compared to three baseline methods, the heuristics SR-RAG uses achieves the best F1 and AUROC over the short-form and closed-set subset of the training set. Building upon this paper's results, future work can further study leveraging more advanced uncertainty quantification methods \citep{wu-etal-2024-synchronous, DBLP:journals/corr/abs-2501-12835} to build more accurate source selection labels.

\begin{table*}[t]
    \centering
    \resizebox{\linewidth}{!}{
    \begin{tabular}{l|cc|cc|cc|cc|cc}
        \toprule
        \multirow{2}{*}{\textbf{Method}} & \multicolumn{2}{c|}{\textbf{ARC\_Easy}} & \multicolumn{2}{c|}{\textbf{NQ}} & \multicolumn{2}{c|}{\textbf{OBQA}} & \multicolumn{2}{c|}{\textbf{Fever}} & \multicolumn{2}{c}{\textbf{Average}} \\
         & \textbf{F1} & \textbf{AUROC} & \textbf{F1} & \textbf{AUROC} & \textbf{F1} & \textbf{AUROC} & \textbf{F1} & \textbf{AUROC} & \textbf{F1} & \textbf{AUROC} \\
        \midrule
        Best Single Likelihood & 0.640 & 0.572 & 0.320 & 0.551 & 0.660 & 0.606 & 0.560 & 0.650 & 0.545 & 0.585 \\
        Best Average Likelihood & 0.730 & \textbf{0.692} & 0.420 & 0.626 & \textbf{0.720} & \textbf{0.643} & 0.630 & 0.703 & 0.630 & 0.666 \\
        Best All Rank & 0.700 & 0.691 & \textbf{0.470} & 0.632 & 0.700 & 0.621 & 0.630 & 0.674 & 0.625 & 0.655 \\
        SR-RAG & \textbf{0.740} & 0.691 & 0.440 & \textbf{0.635} & \textbf{0.720} & 0.635 & \textbf{0.670} & \textbf{0.711} & \textbf{0.643} & \textbf{0.668} \\
        \bottomrule
    \end{tabular}
    }
    \caption{\textbf{Source labeling accuracy of different heuristics.} F1 and AUROC scores for different methods across datasets. Llama-2-7B-Chat is used as the LLM. The best score per column is boldfaced.}
    \label{tab:source_lab_heuristics}
\end{table*}

\begin{figure*}[t]
\centering
\includegraphics[width=0.99\textwidth]{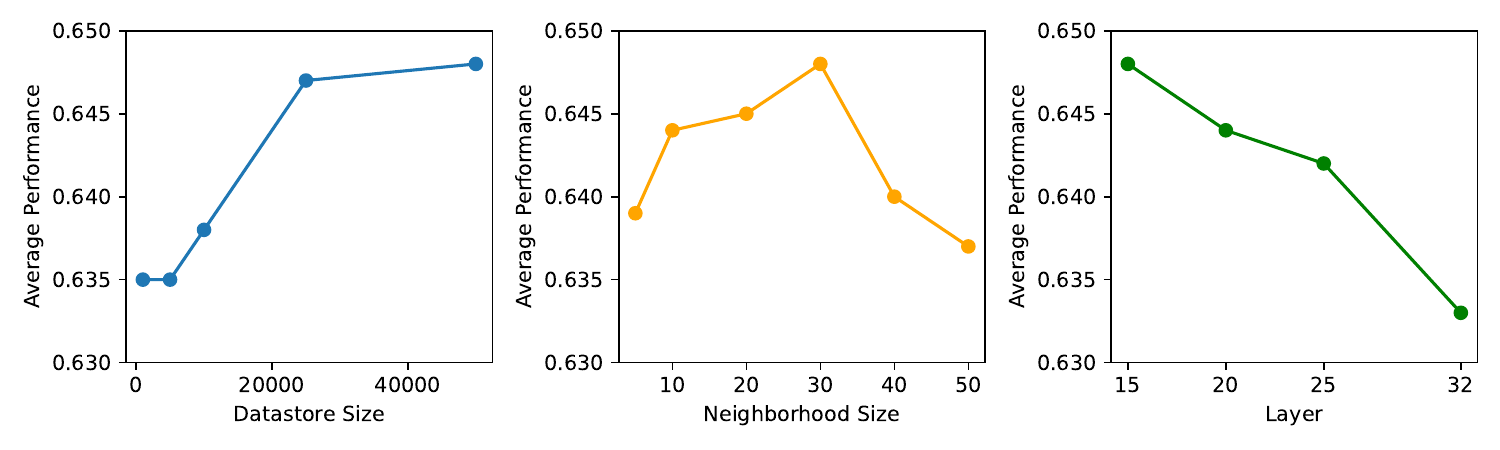}
\caption{\textbf{Hyperparameter settings for the kNN policy datastore with Llama-2-7B-Chat.}} 
\label{fig:knn-hyperparams}
\end{figure*}

\paragraph{Datastore Size} To study the influence of the kNN policy datastore size on the performance of SR-RAG, we randomly sample a subset of the training set (50k samples in total) to construct the datastore. As shown in \Cref{fig:knn-hyperparams} (blue), reducing the kNN policy datastore to 25k (half) has minimal impact. Even with just 1k examples, performance drops only slightly, suggesting potential for lightweight and memory-efficient deployments.

\paragraph{Number of Neighbors} The number of neighbors retrieved from the datastore determines the localness of the focus in the hidden representation space. While task-specific neighborhood size can be tuned, SR-RAG generally works well with 10 to 30 neighboring data points (\Cref{fig:knn-hyperparams} middle), indicating its robustness. However, using a too small (e.g., 5) or a too large (e.g., more than 50) neighborhood harms the performance of SR-RAG.

\paragraph{Layer for Hidden State} We further study the impact of layer selection on SR-RAG's performance in \Cref{fig:knn-hyperparams} part 3 (green). Overall, we observe that using a middle layer's hidden states has better performance over using the last layer's. In the next section, we visualize the hidden state space across LLMs and show that middle layers tend to learn better representations for the model uncertainty as well as task types.

\subsection{Hidden State Space of the Source Reflection Token}
\label{further-analyses-hidden-state-vis}

Does SR-RAG training allow the model to condense its knowledge of tasks, model uncertainty, and source characteristics into the representation of \texttt{<EOQ>}? 

\begin{figure*}[t]
\centering
\includegraphics[width=0.94\textwidth]{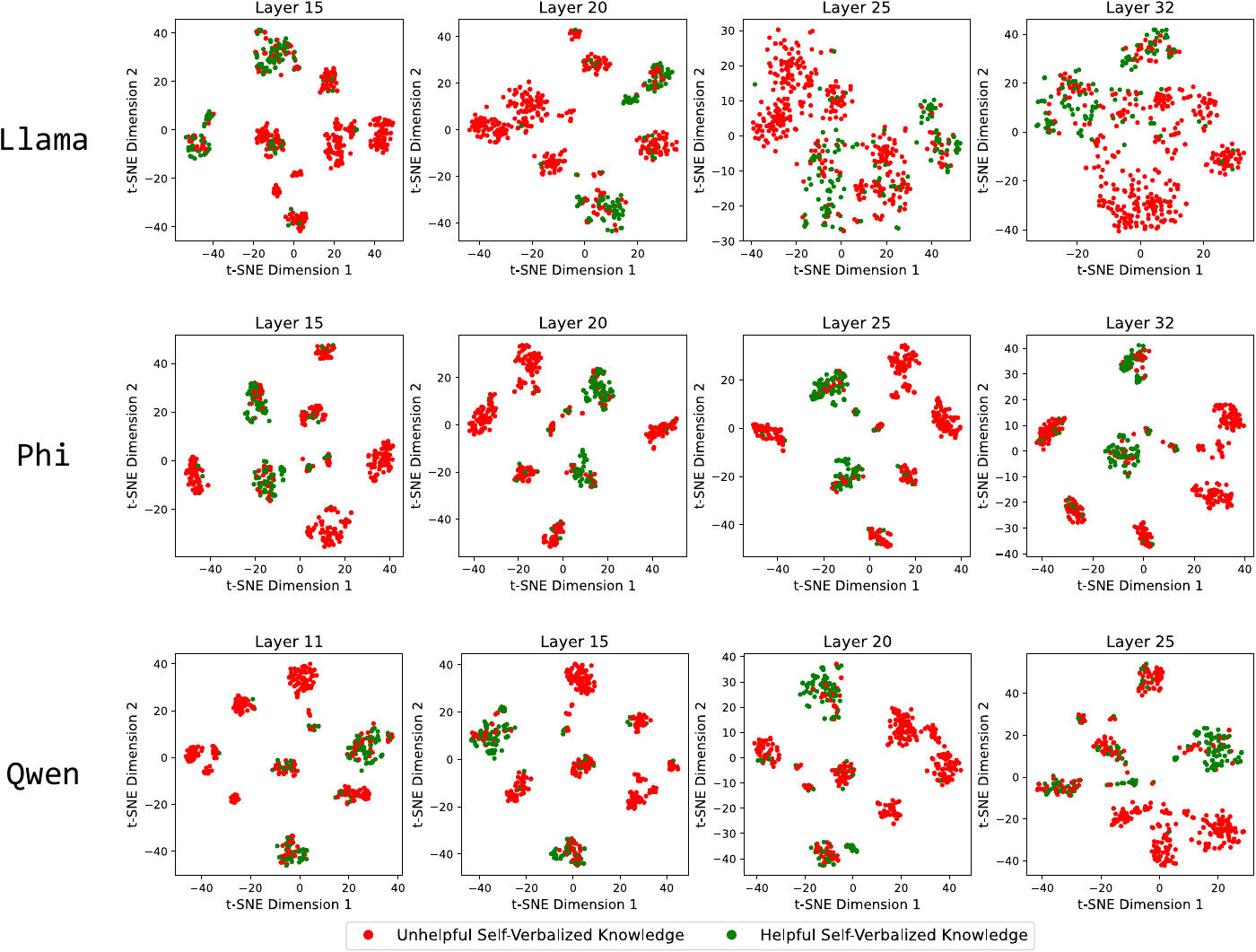}
\caption{\textbf{Hidden State Visualization 1.} Hidden states of \texttt{<EOQ>} across three LLMs on PopQA, visualized using t-SNE. Llama = Llama-2-7B-Chat, Phi = Phi-3.5-Mini-Instruct, and Qwen = Qwen2.5-7B-Instruct.} 
\label{fig:hidden-state-popqa}
\end{figure*}

\paragraph{Uncertainty Encoding in \texttt{<EOQ>}}
In \Cref{fig:hidden-state-popqa}, we visualize \texttt{<EOQ>} hidden states on PopQA using t-SNE. Clear separation emerges between instances where self-verbalized knowledge is helpful (green) and unhelpful (red). Middle layers show stronger clustering than final layers, supporting the use of these representations for kNN-based source selection.

\begin{figure*}[t]
\centering
\includegraphics[width=0.94\textwidth]{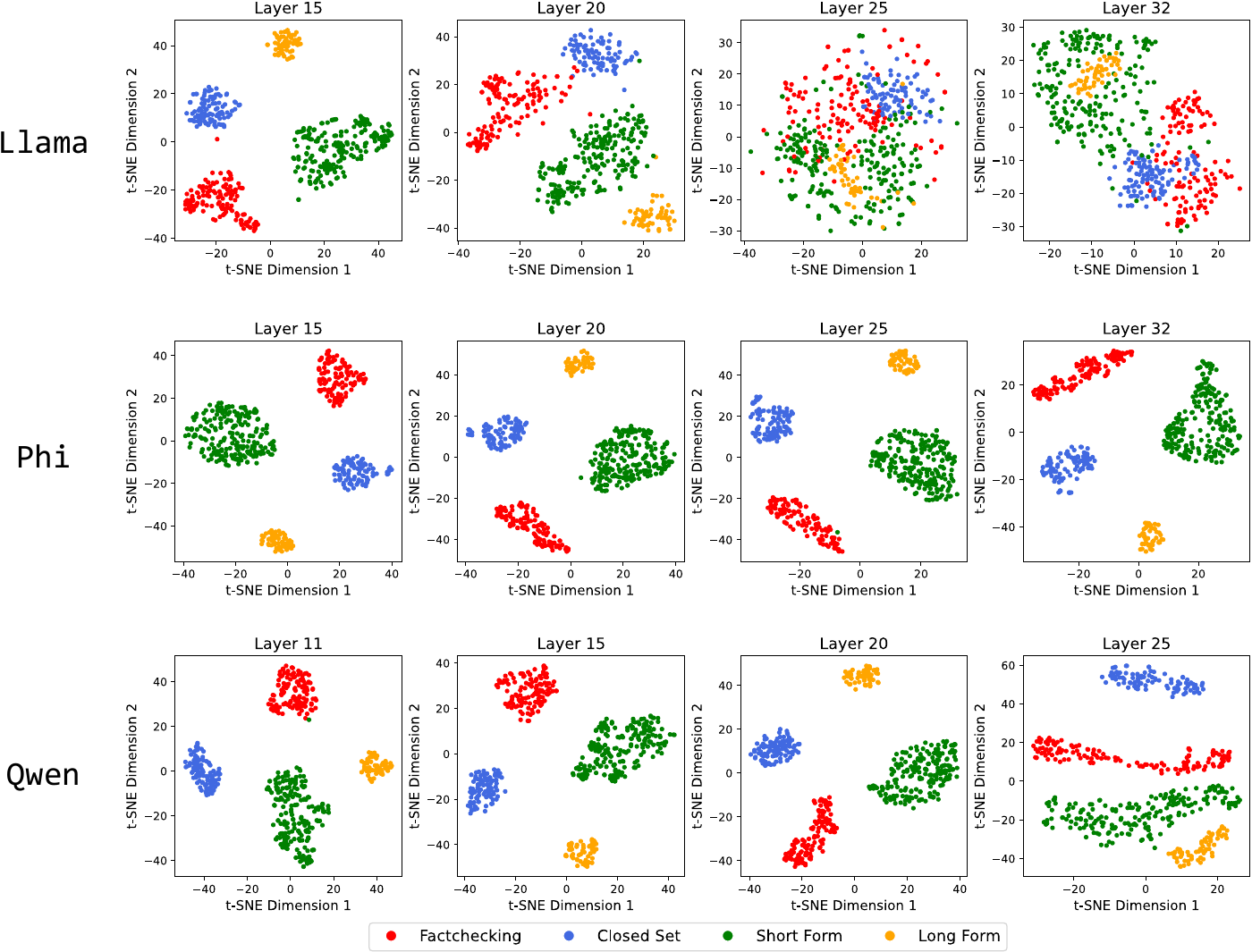}
\caption{\textbf{Hidden State Visualization 2.} Hidden states of \texttt{<EOQ>} across three LLMs on a random sample of 500 data points in the SR-RAG training set, visualized using t-SNE. Llama = Llama-2-7B-Chat, Phi = Phi-3.5-Mini-Instruct, and Qwen = Qwen2.5-7B-Instruct.} 
\label{fig:hidden-state-all-tasks}
\end{figure*}

\paragraph{Implicit Task Clustering} We further randomly sample 500 data points from the SR-RAG train set and visualize their hidden states. 
\Cref{fig:hidden-state-all-tasks} shows that SR-RAG also learns to cluster instances by task type (e.g., fact-checking, closed QA, long-form generation). This suggests that source reflection tokens carry semantically meaningful and task-sensitive representations on top of uncertainty information. These behaviors serve as the foundation of our nearest neighbor-based source selection approach.  

\begin{table*}[ht!]
    \centering
\begin{tabular}{c|cccc|c}
\toprule
 \textbf{Source for $c_{i+}$ and $c_{i-}$} & \textbf{PopQA} & \textbf{TriviaQA} & \textbf{PubHealth} & \textbf{ARC} & \textbf{Average} \\
\midrule
\rowcolor{gray!20} \multicolumn{6}{c}{\textbf{\%Verb $\geq$ RAG $(\uparrow)$}} \\
\midrule
Self-Generated Knowledge & \textbf{33.0\%} & \textbf{56.6\%} & \textbf{73.4\%} & \textbf{63.5\%} & \textbf{56.6\%} \\
Distilled Knowledge & 28.2\% & 55.3\% & 69.7\% & 62.3\% & 53.9\% \\

\midrule
\rowcolor{gray!20} \multicolumn{6}{c}{\textbf{Verbalization Performance $(\uparrow)$}} \\
\midrule
Self-Generated Knowledge & \textbf{0.302} & \textbf{0.572} & \textbf{0.734} & \textbf{0.634} & \textbf{0.561} \\
Distilled Knowledge & 0.282 & 0.553 & 0.697 & 0.606 & 0.535 \\

\bottomrule
\end{tabular}
 \caption{\textbf{Effect of knowledge distillation.} Comparison between training with self-generated knowledge versus knowledge distillation for the verbalization branch in SR-RAG. Llama-2-7B-Chat is fine-tuned either with self-verbalized knowledge or the knowledge verbalized from Llama-3.3-70B-Instruct.}
\label{tab:self-vs-other-knowledge}
\end{table*}

\subsection{Can SR-RAG learn more from self-generated knowledge?}
\label{further-analyses-70b-distillation}

In SR-RAG training, self-verbalized knowledge contexts from the LLM are utilized to fine-tune the model via behavior cloning (stage 1) or preference alignment (stage 2). Alternatively, is it possible that the model can learn to verbalize even higher-quality knowledge from off-policy samples generated by a stronger LLM? We investigate this hypothesis by running GenRead-based knowledge verbalization on Llama-3.3-70B-Instruct and directly use the knowledge to prepare SR-RAG training data with Llama-2-7B-Chat. \Cref{tab:self-vs-other-knowledge} compares the performance of fine-tuning Llama-2-7B-Chat with this data versus using self-generated knowledge for SR-RAG. Interestingly, using self-generated positive and negative knowledge pairs is more beneficial for unlocking the model's ability to generate useful knowledge, outperforming the alternative by 4.9\% higher answer accuracy and 5.0\% higher win rate over retrieval. It is also possible that the ability of the teacher model needs to be close to the fine-tuned model for its knowledge to be useful. We leave this investigation to future work.
\subsection{Discussion: Efficiency and Applicability}
\label{discussion-efficiency-applicability}

In this section, we briefly discuss the efficiency and applicability of SR-RAG.

\paragraph{Training Efficiency}
SR-RAG introduces additional offline computation because source labels are constructed by collecting multiple candidate contexts per source (retrieved passages and self-verbalizations) and scoring them with answer likelihood (\Cref{appendix-training-details}). We view this cost as a one-time post-training expense that can be amortized across deployments, and it requires no extra human annotation or stronger teacher models. In addition, our two-stage recipe mirrors common modern post-training pipelines that combine supervised fine-tuning with preference alignment, and it can be incorporated without substantial changes to standard training infrastructure.

\paragraph{Inference Efficiency}
SR-RAG preserves efficient inference by executing routing, knowledge collection, and answer generation in a single left-to-right pass (\Cref{fig:main-framework}). When the internal source is selected, the model produces one compact knowledge verbalization instead of running multi-sample elicitation; when an external source is selected, decoding pauses only to issue retrieval, and overall latency is largely governed by retrieval frequency (\Cref{fig:system-efficiency-main-text,appendix-latency-formulation}). The kNN policy adds a lightweight computation when the datastore is cached on GPU, and our hyperparameter analyses show that routing quality is not highly sensitive to datastore size, suggesting that compact or compressed datastores can be effective (\Cref{fig:knn-hyperparams}).

\paragraph{Applicability}
SR-RAG is designed for practical routing settings where systems must choose among heterogeneous sources. First, SR-RAG achieves favorable accuracy-efficiency trade-offs without dataset-specific threshold tuning, which reduces the manual effort required to deploy selective retrieval across diverse workloads (\Cref{fig:system-efficiency-main-text}). Second, the policy datastore improves interpretability because it stores explicit source assignments that can be inspected and debugged, rather than relying solely on opaque likelihood thresholds. Third, the same datastore also enables controllability: practitioners can audit and modify routing behavior by editing, filtering, or augmenting stored examples (for instance, enforcing least-privilege routing or overriding problematic regions of the decision boundary). Finally, SR-RAG naturally extends beyond the two-source setup, and our three-source experiments illustrate that the framework can route to specialized corpora (e.g., PubMed) when needed while retaining strong performance on general workloads (\Cref{srrag-extensions}).

\subsection{Qualitative Study}
\label{further-analyses-examples}

In \Cref{qualitative-study-1} and \Cref{qualitative-study-2}, we show two qualitative examples from TriviaQA with the fine-tuned Llama-2-7B-Chat as the backbone LLM. For both examples, we show the LLM's prediction conditioned on both retrieved knowledge and the verbalized knowledge. While the verbalized knowledge is more concise than retrieved paragraphs, it better identifies the correct answers for both questions. This is because the core information (i.e., "first name", "island") only takes up a small portion of the query. As a result, dense retrieval is noise-prone and brings in only distantly relevant information. By contrast, the LLM as a knowledge generator dynamically adjusts its focus, guiding the knowledge generation to include the core information.  As a result, the LLM can correctly answer both questions with self-verbalized knowledge but provides incorrect answers with the low-quality retrieved knowledge. In the source selection process, the LLM directly predicts the source preference $p_{M}(\text{\texttt{\textless s\textgreater}} | q)$. In addition, questions from the train set sharing the same topic and difficulty level are retrieved to form the source preference distribution $p_{D}(\text{\texttt{\textless s\textgreater}} | q)$. In the first example (\Cref{qualitative-study-1}), both distributions prefer the LLM itself as the knowledge source. By contrast, in the second example (\Cref{qualitative-study-2}), $p_{M}(\text{\texttt{\textless s\textgreater}} | q)$ has a high uncertainty and $p_{D}(\text{\texttt{\textless s\textgreater}} | q)$ corrects the source distribution to select the LLM itself as the knowledge source.

\begin{figure*}[ht!]
\small
\centering
\begin{tabular}{p{0.98\linewidth}}
    \hline
    \textbf{Question: } In the book David Copperfield what was Mr. Micawber's first name?  \\
    \textbf{Answer: } Wilkins \\
    \hline
    \textbf{Retrieved Knowledge:} [1] David Copperfield (1999 film) a visit from Mr. Peggotty, back from Australia. He brings news that Emily has made a full recovery and that Mr. Micawber has established himself as a successful magistrate and bank manager. The story closes with the birth of David and Agnes's third child - a girl. Betsey Trotwood's wish finally comes true after nearly 30 years, as David decides that the baby will be christened Betsey Trotwood Copperfield, in honour of her godmother. [2]... (following knowledge chunks omitted) \\
    \textbf{Prediction (with retrieved knowledge):}  \textcolor{red}{John Micawber, Sr.}, is a fictional character in Charles Dickens's 1850 novel David Copperfield. He is the father of David Copperfield's godfather, Mr. Micawber, Jr., and is portrayed as a kind and gentle man who is often in financial difficulties. Mr. Micawber's first name is Charles, not John, as is often mistaken.    \\
    \hline
    \textbf{Verbalized Knowledge:} David Copperfield is a novel by Charles Dickens, published in 1849-50. The novel follows the life of David Copperfield, a young man who grows up in Victorian England. Mr. Micawber is a character in the novel, a kind and well-meaning man who is a friend of David's father. Mr. Micawber's first name is Wilkins. \\
    \textbf{Prediction (with verbalized knowledge):} \textcolor{teal}{Wilkins Micawber.} Mr. Micawber is a character in Charles Dickens 's novel David Copperfield. He is a kind and well - meaning friend of David 's father, Mr. Copperfield. Mr. Micawber is known for his optimistic outlook on life and his tendency to get into financial difficulties.    \\
    \hline
    \textbf{Neighbor Questions: }   \\
    - What was Mr. Smith's first name in Mr. Smith goes to Washington? (\textcolor{teal}{\texttt{<Self>}} $>$ \textcolor{red}{\texttt{<Wiki>}}) \\
    - Who was the original singer of Knocking on Heaven's Door? (\textcolor{teal}{\texttt{<Self>}} $>$ \textcolor{red}{\texttt{<Wiki>}})  \\
    - When did sir Arthur Conan Doyle wrote his first book? (\textcolor{red}{\texttt{<Wiki>}} $>$ \textcolor{teal}{\texttt{<Self>}}) \\
    ... (27 questions omitted)\\
    $p_{D}(\text{\texttt{\textless s\textgreater}} | q)$: P(\textcolor{red}{\texttt{<Wiki>}}) = 9/30 = 0.3, P(\textcolor{teal}{\texttt{<Self>}}) = 21/30 = 0.7.  \\
    $p_M(\text{\texttt{\textless s\textgreater}} | q)$: P(\textcolor{red}{\texttt{<Wiki>}}) = 0.146, P(\textcolor{teal}{\texttt{<Self>}}) = 0.854. \\
    \textbf{Final Source Decision: } \textcolor{teal}{\texttt{<Self>}} \\
    \hline
\end{tabular}
\caption{\textbf{Qualitative Example 1.} Example of SR-RAG inference where $p_{D}(\text{\texttt{\textless s\textgreater}} | q)$ agrees with $p_M(\text{\texttt{\textless s\textgreater}} | q)$. As the model itself is selected as the knowledge source, the retrieval will be abstained. The retrieved knowledge and the corresponding prediction are shown only for illustration. }
\label{qualitative-study-1}
\end{figure*}

\begin{figure*}[ht!]
\small
\centering
\begin{tabular}{p{0.98\linewidth}}
    \hline
    \textbf{Question: } From 1791 to 1803, Toussaint L’Ouverture led a revolt against France on which island?   \\
    \textbf{Answer: } Haiti \\
    \hline
    \textbf{Retrieved Knowledge:} [1] Saint-Domingue expedition slave revolt that led to the abolition of slavery in 1793 by the civil commissioners Sonthonax and Polverel, in a decision endorsed and spread to all the French colonies by the National Convention 6 months later. Toussaint Louverture, a black former slave who had been made Governor by France, re-established peace, fought off Spanish and British attempts to capture the island, and reestablished prosperity by daring measures. [2]... (following knowledge chunks omitted) \\
    \textbf{Prediction (with retrieved knowledge):}  \textcolor{red}{Saint-Pierre et Miquelon (Saint-Pierre and Miquelon), a French overseas collectivity in the North American Atlantic Ocean.}    \\
    \hline
    \textbf{Verbalized Knowledge:} Toussaint L'Ouverture was a Haitian leader who led a successful revolt against French colonial rule in Haiti from 1791 to 1803. L'Ouverture was a former slave who became a prominent figure in the Haitian Revolution, which was a series of uprisings and battles fought between Haitian slaves and their French colonial oppressors. After years of fighting, L'Ouverture and his forces were able to secure Haiti's independence from France on January 1, 1804. \\
    \textbf{Prediction (with verbalized knowledge):} \textcolor{teal}{Haiti (Saint-Domingue).}  \\
    \hline
    \textbf{Neighbor Questions: }   \\
    - The battle of Hastings in 1066 was fought in which country? (\textcolor{teal}{\texttt{<Self>}} $>$ \textcolor{red}{\texttt{<Wiki>}}) \\
    - When Belgium declared its independence in 1830 it broke away from \_\_\_ control? (\textcolor{teal}{\texttt{<Self>}} $>$ \textcolor{red}{\texttt{<Wiki>}})  \\
    - When did the French come to the new world? (\textcolor{teal}{\texttt{<Self>}} $>$ \textcolor{red}{\texttt{<Wiki>}}) \\
    ... (27 questions omitted)\\
    $p_{D}(\text{\texttt{\textless s\textgreater}} | q)$: P(\textcolor{red}{\texttt{<Wiki>}}) = 24/30 = 0.8, P(\textcolor{teal}{\texttt{<Self>}}) = 6/30 = 0.2.  \\
    $p_M(\text{\texttt{\textless s\textgreater}} | q)$: P(\textcolor{red}{\texttt{<Wiki>}}) = 0.576, P(\textcolor{teal}{\texttt{<Self>}}) = 0.424. \\
    \textbf{Final Source Decision: } \textcolor{teal}{\texttt{<Self>}} \\
    \hline
\end{tabular}
\caption{\textbf{Qualitative Example 2.} Example of SR-RAG inference where $p_{D}(\text{\texttt{\textless s\textgreater}} | q)$ corrects the source selection from $p_M(\text{\texttt{\textless s\textgreater}} | q)$. The model itself is selected as the knowledge source.}
\label{qualitative-study-2}
\end{figure*}

\clearpage

\end{document}